\documentclass{article}

\usepackage[preprint]{neurips_2026}

\usepackage{amsmath}  
\usepackage{pdfpages}
\usepackage{amssymb}  
\usepackage{mathtools}
\usepackage{subcaption}
\usepackage[utf8]{inputenc} 
\usepackage[T1]{fontenc}    
\usepackage{hyperref}       
\usepackage{url}            
\usepackage{booktabs}       
\usepackage{amsfonts}       
\usepackage{nicefrac}       
\usepackage{microtype}      
\usepackage{xcolor}         
\usepackage{multirow}
\usepackage{graphicx}
\usepackage[title]{appendix}
\usepackage{float}

\title{Learning Residual Kinematic Corrections for Continuous Neural Decoding via Reinforcement Learning}

\author{%
  \textbf{Jiamian Li}$^1$, \textbf{Niall McShane}$^2$, \textbf{Attila Korik}$^3$, \textbf{Naomi du Bois}$^3$, \textbf{Karl McCreadie}$^2$, \\
  \textbf{Leen Jabban}$^4$, \textbf{Benjamin Metcalfe}$^4$, \textbf{Özgür Şimşek}$^1$, \textbf{Damien Coyle}$^3$ \\[0.15in]
  $^1$Department of Computer Science, University of Bath, Bath, BA2 7AY \\ \texttt{\{jl4668, os435\}@bath.ac.uk} \\[0.05in]
  $^2$Intelligent Systems Research Centre, University of Ulster, BT48 7JL \\ \texttt{\{N.McShane, k.mccreadie\}@ulster.ac.uk} \\[0.05in]
  $^3$Bath Institute for the Augmented Human, University of Bath, Bath, BA2 7AY \\ \texttt{\{ak3825, ndb36, dhc30\}@bath.ac.uk} \\[0.05in]
  $^4$Department of Electronic and Electrical Engineering, University of Bath, Bath, BA2 7AY \\ \texttt{\{lj386, bwm23\}@bath.ac.uk} \\
}

\begin{document}

\maketitle

\begin{abstract}

Decoding continuous three-dimensional (3D) motor imagery (MI) using non-invasive electroencephalography (EEG)-based brain--computer interfaces (BCIs) remains challenging due to signal variability and residual decoding errors. Deep learning architectures such as convolutional neural network--long short-term memory (CNN--LSTM) models can capture spatial and temporal dynamics for continuous kinematic decoding; however, systematic residual errors persist in predicted trajectories. We propose a two-stage decoding framework that applies reinforcement learning (RL) to perform residual kinematic correction on the outputs of a CNN--LSTM decoder (CNN--LSTM--RL). The RL agent is trained offline without direct EEG input and instead operates on predicted kinematic trajectories to optimize movement accuracy relative to target trajectories. This design enables targeted correction of systematic decoder errors while preserving the primary neural decoding pipeline. The proposed framework was evaluated offline using data from ten participants across ten sessions of online continuous 3D motor imagery, with feedback provided in both 2D and immersive virtual reality (VR) environments. Decoding performance was quantified using Pearson correlation coefficients ($r$) and Root Mean Square Errors (RMSE) along the $x, y$, and $z$ axes. Compared to CNN--LSTM applied alone, CNN--LSTM--RL improved the mean correlation from $0.5076$ to $0.7181$ ($p = 0.0005$) in 2D and from $0.6420$ to $0.7780$ ($p = 0.0059$) in VR, with relative gains of $41.5\%$ and $21.2\%$, respectively. Correspondingly, RMSE was reduced from $0.0890$ to $0.0532$ (2D, $p < 0.0001$) and from $0.0714$ to $0.0441$ (VR, $p < 0.0001$), representing relative reductions of $40.2\%$ and $38.2\%$. These findings demonstrate that this scalable framework enhances 3D BCI MI decoding by correcting kinematic errors via offline residual RL without extra neural data, advancing neurorehabilitation, prosthetics, and virtual interaction.
\end{abstract}

\section{Introduction}
Brain-computer interfaces (BCIs) have demonstrated remarkable success across a wide range of domains, including cognitive and physical rehabilitation \citep{millan2010combining} \citep{prasad2010applying} \citep{cervera2018brain}, augmentative communication \citep{willett2021high} \citep{willett2023high} \citep{anumanchipalli2019speech}, assessing consciousness \citep{du2026advancing}, entertainment \citep{marshall2013games} \citep{van2013experiencing}, and robotic control \citep{hochberg2012reach} \citep{ajiboye2017restoration} \citep{edelman2019noninvasive}, by establishing a direct signaling pathway between neural activity and external devices. Non-invasive BCIs have become increasingly popular in the BCI field, as they acquire neural signals, which  are subsequently decoded in real-time to execute user-intended commands, via sensors mounted on the human scalp and mitigate the clinical risks associated with invasive approaches that require the surgical implantation of intracortical electrodes, despite the latter offering higher resolution neural interfacing \citep{lebedev2006brain}. Electroencephalography (EEG) remains a predominant non-invasive modality due to its superior temporal resolution and cost-efficiency relative to techniques such as functional near-infrared spectroscopy (fNIRS) \citep{lotte2018review} \citep{abiri2019comprehensive}.

\noindent Artificial intelligence (AI)–inspired models have proven highly effective in decoding continuous three-dimensional (3D) limb motion trajectories and motion imagery from EEG signals \citep{bradberry2010reconstructing} \citep{chen2022continuous} \citep{wang2023comprehensive} \citep{xu2021decoding} \citep{korik2025decoding}. Recent deep learning approaches, specifically hybrid convolutional neural network and long short-term memory (CNN--LSTM) architectures \citep{mc2023decoding} \citep{mcshane2026embodied}, have demonstrated superior decoding accuracy compared to multiple linear regression models \citep{mc2022online}.

\noindent Reinforcement Learning (RL) is another AI paradigm that formulates problems as sequential decision-making processes and has demonstrated strong effectiveness in complex control tasks \citep{sutton1998reinforcement} \citep{kober2013reinforcement} \citep{singh2022reinforcement} \citep{tang2025deep} \citep{sharifi2025reinforcement}. Unlike supervised deep learning methods, RL does not rely on labeled data; instead, it learns optimal policies through interaction with the environment. Many studies have combined RL with EEG-based brain–computer interfaces using error-related potentials (ErrPs), which are neural signatures of error awareness in EEG, as reward signals \citep{kim2017intrinsic} \citep{wang2022error} \citep{xavier2025hybrid}. A second line of work focuses on adaptive BCI decoding, where RL operates as a post-decoding optimization layer that improves performance through continuous interaction with the user \citep{xavier2025hybrid} \citep{xu2026brain}. Another direction focuses on human preference learning and human-in-the-loop reinforcement learning. In this paradigm, neural signals are leveraged as implicit feedback to guide policy learning \citep{agarwal2020human} \citep{xu2021accelerating} \citep{zhu2025eeg}. Residual Reinforcement Learning \citep{silver2018residual} \citep{johannink2019residual} \citep{alakuijala2021residual} \citep{davchev2022residual} is a widely used method in the field of robotics, designed to output a residual action and correct prediction errors from a baseline policy. However, its efficacy in BCI interaction remains unexplored.

Although existing studies have integrated RL into BCI decoding, they primarily rely on direct EEG features as reward signals. However, the signal variability, non-stationarity, and residual decoding errors of non-invasive EEG often results in the RL agent and the BCI decoder receiving unstable guidance\citep{hameed2025enhancing} \citep{xu2026brain}. Furthermore, most RL-BCI approaches require time-consuming online updates during live interaction, which is often impractical. To address these gaps and the performance limitations of CNN--LSTM BCI decoders on the task of continuous 3D motion imagery decoding,  we extend the work of \citep{mcshane2026embodied} and propose an offline residual reinforcement learning framework (CNN--LSTM--RL) that decouples the correction process from direct neural noise and separates data collection from RL training. Our key contributions are as follows: (1) we introduce a novel two-stage architecture that utilizes a frozen CNN--LTSM decoder to extract spatial-temporal features while allowing the offline RL agent to be trained independently of live user interaction, (2) we enhance decoding robustness and reduce systematic residual errors by enabling the RL agent to perform residual kinematic corrections based solely on the decoder's output and temporal context, a design that effectively isolates the correction process from the direct interference of high-noise EEG signals, achieving relative improvements in mean correlation of up to $41.5\%$ and mean RMSE reductions of up to $40.2\%$.

\section{Methodology}
\subsection{Dataset and Baseline BCI Decoder}
Ten right-handed adult participants took part in the experiment (ethics statement included in Appendix \ref{app:ethics}), completing ten sessions across different days under alternating visual feedback, 2D screen and immersive VR. Across sessions, participants engaged in upper-limb reaching task and imagined reaching tasks to 4 targets. Each session comprised eight runs (four executed and four imagined), with two blocks of 16 trials per run. Within each block, four targets (see Appendix \ref{app:reach targets}) were presented in random order, each appearing four times. This design produces 256 trials per session and a total of 25,600 trials across the complete dataset.

The experimental sequence followed a trial-based structure, where each block was self-initiated and preceded by a 30-second countdown. Individual trials, spanning 7.6 s, consisted of four sequential functional stages: a 2.5 s rest phase for baseline fixation and neural stability, followed by a 1.6 s Indication Phase where a directional pre-cue line pointed toward the 3D target. The subsequent 2.5 s target phase was initiated by a 440 Hz auditory tone (640 ms) and featured a 1.5$\times$ target expansion alongside animated trajectory guidance to facilitate motor imagery. Each trial concluded with a 1.0 s Reset Phase, signaled by a secondary auditory cue, which prompted the return of the virtual hand to its initial starting position. The movement-onset segment was extracted solely from the runs within the target phase to train both the BCI decoder and the RL agent. The segments were temporally aligned to the trajectory from the resting position toward the target hold, excluding the return movement during the reset phase, and maintained a fixed duration of 2 s across all trials.

EEG neural activity was captured using a wireless g.Nautilus system (g.tec) with 32 g.SCARABEO active electrodes. While a comprehensive 32-channel montage was recorded (including C3, C4, AF3, T7, etc.), online decoding and decoder training were streamlined using a targeted 17-channel subset (F3, FZ, F4, FC5, FC1, FC2, FC6, C3, CZ, C4, CP5, CP1, CP2, CP6, P3, PZ, and P4) centered over the sensorimotor cortex. Concurrently, kinematic trajectories were sampled at $60$~Hz via a wrist-mounted Vive tracker. This system provided continuous three-dimensional Cartesian coordinates ($x, y, z$), which were subsequently subjected to numerical differentiation to derive instantaneous velocity estimates for each time step.

The BCI decoder was developed with an Event-Related Spectral Perturbation (ERSP)-based \citep{makeig1993auditory} CNN–LSTM architecture to decode continuous 3D limb velocity from EEG signals. ERSP provided time–frequency representations of motor-related neural activity by quantifying spectral power changes relative to a baseline. For each of the 17 EEG channels, 2D ERSP images (40 × 40 pixels) spanning 0–40 Hz and a time window with 16 ms as observation steps from 640 ms pre-movement through execution were generated and stacked into a channel-wise input volume. The CNN component, consisting of three convolutional layers with 3 × 3 kernels and 32 filters in the first layer with max-pooling and dropout to reduce overfitting, learned spatial features from these ERSP representations. Extracted features were then fed into two stacked LSTM layers, with 50 hidden units in the first LSTM layer and a dropout rate of 0.25 applied between recurrent layers, to capture temporal dependencies across sequential windows and maintain contextual information. The network produced continuous predictions of x-, y-, and z-axis velocities through a dense output layer with tanh activations in hidden layers.

\subsection{Reinforcement Learning Problem formulation}
The RL problem can be formulated as a discrete-time Partially Observable Markov Decision Process (POMDP) for residual control, defined by the tuple $\mathcal{M} = \langle \mathcal{S}, \mathcal{A}, \mathcal{T}, \mathcal{R}, \Omega, \mathcal{O}, \gamma \rangle$, where $\mathcal{S}$ and $\mathcal{A}$ denote the state and action spaces, $\mathcal{R}$ the reward function, $\Omega$ the observation space, $\mathcal{O}$ the observation model, and $\gamma \in [0, 1)$ the discount factor. The true cognitive state of the user is hidden, and the reinforcement learning agent must optimize a residual policy based solely on the outputs of a pre-trained base decoder. Let $s_t \in \mathcal{S}$ denote the true, hidden state of the system at time step $t$. It comprises the user's unobservable brain signals $EEG_t$, the current BCI decoded velocities $v^{BCI}_t$, the inaccessible target velocity trajectory $v^*_t$ (used for reward computation but not included in the agent’s observations), the RL agent's final corrected velocities $v_{t-1}$ at the last timestep, and the current time step $t$. Thus, $s_t = [EEG_t, v^{BCI}_t, v^*_t, v_{t-1}, t]$. The action $a_t \in \mathcal{A}$ is defined as the continuous residual velocity $v^{res}_{t}$ generated by the RL policy. The final action applied to the environment is $v_{t} = v^{BCI}_t + v^{res}_{t}$. The agent receives a partial observation $o_t \in \Omega$ determined by the observation model $\mathcal{O}(o_t \mid s_t)$. The observation is defined as $o_t = [v^{BCI}_t, v_{t-1}, t]$. $\mathcal{T}$ is the state transition function, $s_{t+1} = f(s_t, a_t)$, where $f$ defines the system transition, governing the progression of each state element. The reward function $\mathcal{R}(s_t, a_t)$ is designed to minimize the error between corrected and target trajectory. It is defined as: $r_t = -d(v_{t}, v^*_{t})$, where $d(\cdot, \cdot)$ is a Euclidean distance metric between the corrected velocities $v_{t}$ and the target velocities $v^*_t$.

\subsection{Soft Actor-Critic}
We use the off-policy algorithm Soft Actor-Critic (SAC) \citep{haarnoja2018soft} for RL training, which optimizes for both expected return and policy entropy. The objective function is defined as: \begin{equation}
J(\pi) = \sum_{t=0}^{T} \mathbb{E}_{(\mathbf{s}_t, \mathbf{a}_t) \sim \pi} \left[ r(\mathbf{s}_t, \mathbf{a}_t) + \alpha \mathcal{H}(\pi(\cdot | \mathbf{s}_t)), \right]
\end{equation}
where $\alpha$ is the temperature parameter controlling the exploration-exploitation trade-off.

To optimize this objective, SAC alternates between a soft policy evaluation and a soft policy improvement. The soft state-value function $V^{\pi}(\mathbf{s}_t)$ and the soft action-value function $Q^{\pi}(\mathbf{s}_t, \mathbf{a}_t)$ are related through the following soft Bellman equation:
\begin{equation}
V^{\pi}(s_t) = \mathbb{E}_{a_t \sim \pi} [Q^{\pi}(s_t, a_t)] + \alpha H(\pi(\cdot | s_t).)
\end{equation}
Accordingly, the Bellman equation for the soft action-value function $Q^\pi$ is expressed as:
\begin{align}
Q^{\pi}(s_t, a_t) &= \mathbb{E}_{s_{t+1} \sim P, a_{t+1} \sim \pi} [R(s_t, a_t, s_{t+1}) + \gamma (Q^{\pi}(s_{t+1}, a_{t+1}) + \alpha H(\pi(\cdot | s_{t+1})))] \nonumber \\
&= \mathbb{E}_{s_{t+1} \sim P} [R(s_t, a_t, s_{t+1}) + \gamma V^{\pi}(s_{t+1}),]
\end{align}
where $\gamma$ is the discount factor and $P$ represents the state transition probability distribution. The actor is optimized to maximize both the expected return and the entropy of the policy:
\begin{equation}
J (\pi)= \mathbb{E}_{s_t \sim \mathcal{D}, a_t \sim \pi}
\left[ Q(s_t, a_t) - \alpha \log \pi(a_t | s_t). \right]
\end{equation}
To allow gradient backpropagation, the policy uses the reparameterization trick. The action is expressed as a deterministic function of the state and a Gaussian noise vector $\epsilon_t$:
\begin{equation}
a_t =  \text{tanh}(\mu(\mathbf{s}_t) + \sigma(\mathbf{s}_t) \odot \epsilon_t,)
\end{equation}
where $\mu(\mathbf{s}_t)$ and $\sigma(\mathbf{s}_t)$ denote the state-dependent mean and standard deviation predicted by the policy network, respectively. This transformation ensures that the policy $\pi$ is differentiable.

For implementations, the final action output need to be bounded within a specific range. As the base BCI decoder may require varying degrees of correction across different dimensions (x, y or z), we apply dimension specific scaling factors to the RL agent's actions. This extends the original SAC formulation by allowing each action dimension to be scaled independently. Because the policy $\pi(\mathbf{a}_t | \mathbf{s}_t)$ is constructed by applying a $\text{tanh}$ transformation to a Gaussian distribution followed by using different scales, the log-probability of the action must be corrected. We let $\mathbf{u}_t$ be the sample from the Gaussian distribution and $\mathbf{a}_t = \text{scale} \cdot \text{tanh}(\mathbf{u}_t)$ be the bounded action. The log-density is given by:
\begin{equation}
\log \pi(\mathbf{a}_t | \mathbf{s}_t)
=
\log \mathcal{N}(u_t \mid {\mu}(s_t), {\sigma}^2(s_t))
- \sum_{i=1}^{D} \log \left( 1 - \tanh^2(u_{i,t}) \right)
- \sum_{i=1}^{D} \log(\text{scale}_i)
\end{equation}

\noindent where the first term corresponds to the log-density of the Gaussian distribution before the $\tanh$ squashing, and $u_{i,t}$ denotes the $i$-th element of $\mathbf{u}_t$.

\subsection{Two-Stage Framework}
\begin{figure}[htbp]
    \centering
    \includegraphics[width=\linewidth]{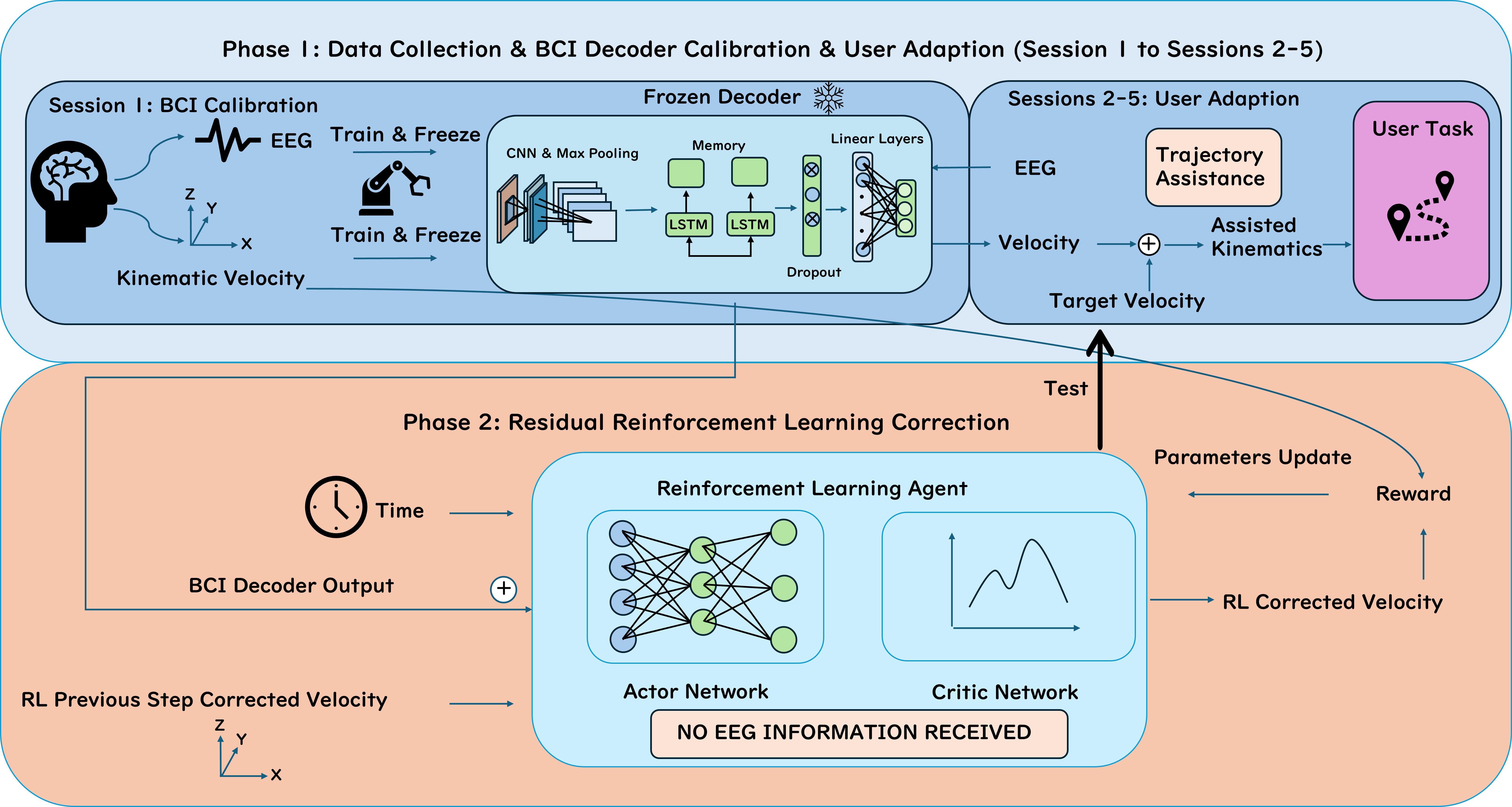}
    \caption{The architecture of our proposed two-stage framework.} 
    \label{fig:two_stage_framework}
\end{figure}
Our architecture consists of two stages, shown in Figure \ref{fig:two_stage_framework}. In the first stage, the baseline CNN--LSTM decoder is calibrated and data are collected. The first session under both 2D and VR feedback is used to calibrate the decoder offline, with separate models trained for each feedback modality and each participant. The calibrated decoder is then frozen and deployed for online data collection and user adaption in the subsequent four sessions, during which trajectory assistance blends decoded BCI trajectory with the target trajectory. In the second stage, an offline reinforcement learning agent is trained on the first session to apply residual correction to the decoder's kinematic output and subsequently evaluated across all remaining sessions. Similar to the decoder, separate RL agents are trained for each feedback modality and each participant. The agent receives the decoder output, normalized time, and the previously corrected velocity as observations, without direct access to EEG signals and target velocities, and outputs a residual action that is added to the baseline decoder output to produce the final corrected velocity.
\section{Experiments}
\subsection{Hyperparameters Optimization and Data Processing}
We use Optuna\citep{optuna_2019} for RL hyperparameters optimization with 200 searching trials, aiming to maximize the correlation coefficient between corrected and target trajectories. The RL agent’s actor and critic networks are implemented using linear layers. The tuned parameters, which include actor learning rate, critic learning rate, hidden layer dimension, batch size, reward discount factor, SAC target smoothing coefficient, scale of residual actions in three dimensions, number of actor layers and number of critic layers, and their range are shown in Appendix \ref{app:hyperparameters searching}. The optimization strategy is the 5-fold cross-validation approach, which used 80\% of the data for training and 20\% of the data for validation, where the average performance across the five folds served as the objective metric for Optuna’s search process; once the best parameters were selected, the model was retrained on the combined training and validation sets and assessed against the four test sessions (S2--S5). The dataset was normalized to the range [0, 0.3] using trial-specific min-max normalization. Details of experiments compute resources are provided in Appendix \ref{app:compute resources}.

\subsection{Evaluation Metric}
The correction accuracy is evaluated using Pearson correlation coefficient and root mean square errors (RMSE). For the $i$-th trial and spatial dimension $d$, the Pearson correlation coefficient (r) between the decoded trajectory and the target trajectory is defined as
\begin{equation}
r_{i,d} =
\frac{
\sum_{t=1}^{T_i} (X_{t,d,i} - \bar{X}_{d,i})(Y_{t,d,i} - \bar{Y}_{d,i})
}{
\sqrt{\sum_{t=1}^{T_i} (X_{t,d,i} - \bar{X}_{d,i})^2}
\sqrt{\sum_{t=1}^{T_i} (Y_{t,d,i} - \bar{Y}_{d,i})^2}
+ \epsilon
}.
\end{equation}
The correlation is then averaged across trials and spatial dimensions as
$
\bar{r}_d = \frac{1}{N} \sum_{i=1}^{N} r_{i,d}
$
and
$
r_{\text{overall}} = \frac{1}{D} \sum_{d=1}^{D} \bar{r}_d,
$ where $i$ indexes trials ($i = 1, \dots, N$), $d$ indexes spatial dimensions ($d = 1, \dots, D$), and $t$ indexes time steps. $T_i$ denotes the number of samples in trial $i$. $X_{t,d,i}$ and $Y_{t,d,i}$ represent the decoded and target points, respectively, while $\bar{X}_{d,i}$ and $\bar{Y}_{d,i}$ denote their temporal means. $N$ is the total number of trials, $D$ is the number of spatial dimensions, and $\epsilon$ is a small constant added to avoid division by zero.

\vspace{\baselineskip}

\noindent For the $i$-th trial and spatial dimension $d$, RMSE between the decoded trajectory and the target trajectory is defined as:
\begin{equation}
\text{RMSE}_{i,d} = \sqrt{\frac{1}{T_i} \sum_{t=1}^{T_i} (X_{t,d,i} - Y_{t,d,i})^2}.
\end{equation}
The RMSE is then averaged across trials and spatial dimensions as 
$\overline{\text{RMSE}}_d = \frac{1}{N} \sum_{i=1}^{N} \text{RMSE}_{i,d}$ 
and 
$\overline{\text{RMSE}}_{\text{overall}} = \frac{1}{D} \sum_{d=1}^{D} \overline{\text{RMSE}}_d$.

\subsection{Evaluating Strategies}
There are several strategies for evaluation. Fixed Decoder Generalisation (FDG) represents true online BCI performance, where the CNN-LSTM decoder trained on Session 1 was deployed unchanged across Sessions 1–5. Performance variations under FDG therefore reflect user learning and adaptation. Sequential Adaptive Training (SAT) is an offline evaluation protocol in which a model is trained on one session (the session preceding the next session) and tested on the subsequent session (e.g., S1→2, S2→3). SAT captures the combined effects of decoder updating and user adaptation across sessions. Within-Session Reconstruction (WSR) provides an upper-bound estimate of decoder capability by training and testing on data from the same session. Similar to FDG, RL strategy Fixed Reinforcement Learning (FRL) only uses data from Session 1 to train the agent and then keeps the trained agent frozen tested on all subsequent sessions.

\section{Results}
Beyond reporting the overall performance of FRL using the Pearson correlation coefficient and RMSE, we provide a detailed statistical evaluation based on the performance of BCI decoder and RL strategies across 10 participants. To ensure statistical robustness, the performance metrics for each participant were first averaged across all trials to yield a single representative value per strategy. Because 2D screen feedback and VR spatial feedback represent distinct feedback environments, statistical analyses were conducted separately within each feedback condition. These stratified analyses were intended to assess decoder strategy performance differences within each feedback environment, rather than to test for a formal interaction effect between the feedback condition and the decoder strategy. For each feedback condition and performance metric, an omnibus repeated-measures comparison was first conducted across decoder strategies. Depending on the distributional characteristics of the paired differences, either a one-way repeated-measures ANOVA \citep{girden1992anova} or Friedman test \citep{friedman1937use} was used to assess overall strategy effects. Significant omnibus effects were followed by planned pairwise comparisons between FRL and each alternative strategy. Paired t-tests were used when the paired differences satisfied normality and symmetry assumptions, whereas Wilcoxon signed-rank tests \citep{wilcoxon1992individual} were used when these assumptions were not met or when a consistent non-parametric approach was required within a metric. Several key statistical variables were employed to analyze these differences: $p$-values were used to determine the statistical significance, which was assessed using an alpha level of $\alpha = 0.05$ of the performance improvements, while Cohen’s $d$\citep{cohen2013statistical} and standardized Wilcoxon effect size $r_z$ were employed to quantify the standardized magnitude of these effects. To account for the multiple comparisons involved within each metric and feedback type, the Holm-Bonferroni correction \citep{holm1979simple} was applied to adjust the raw $p$-values. The prerequisite analysis for these tests, including the validation of normality and symmetry for the paired differences, is detailed in Appendix \ref{app:statistical_analysis} to justify the selection of parametric or non-parametric methods for each specific group-level comparison.

\subsection{Pearson Correlation Coefficient Comparison}

\begin{figure}[htbp]
    \centering
    \begin{subfigure}[b]{0.48\linewidth}
        \centering
        \includegraphics[width=\linewidth]{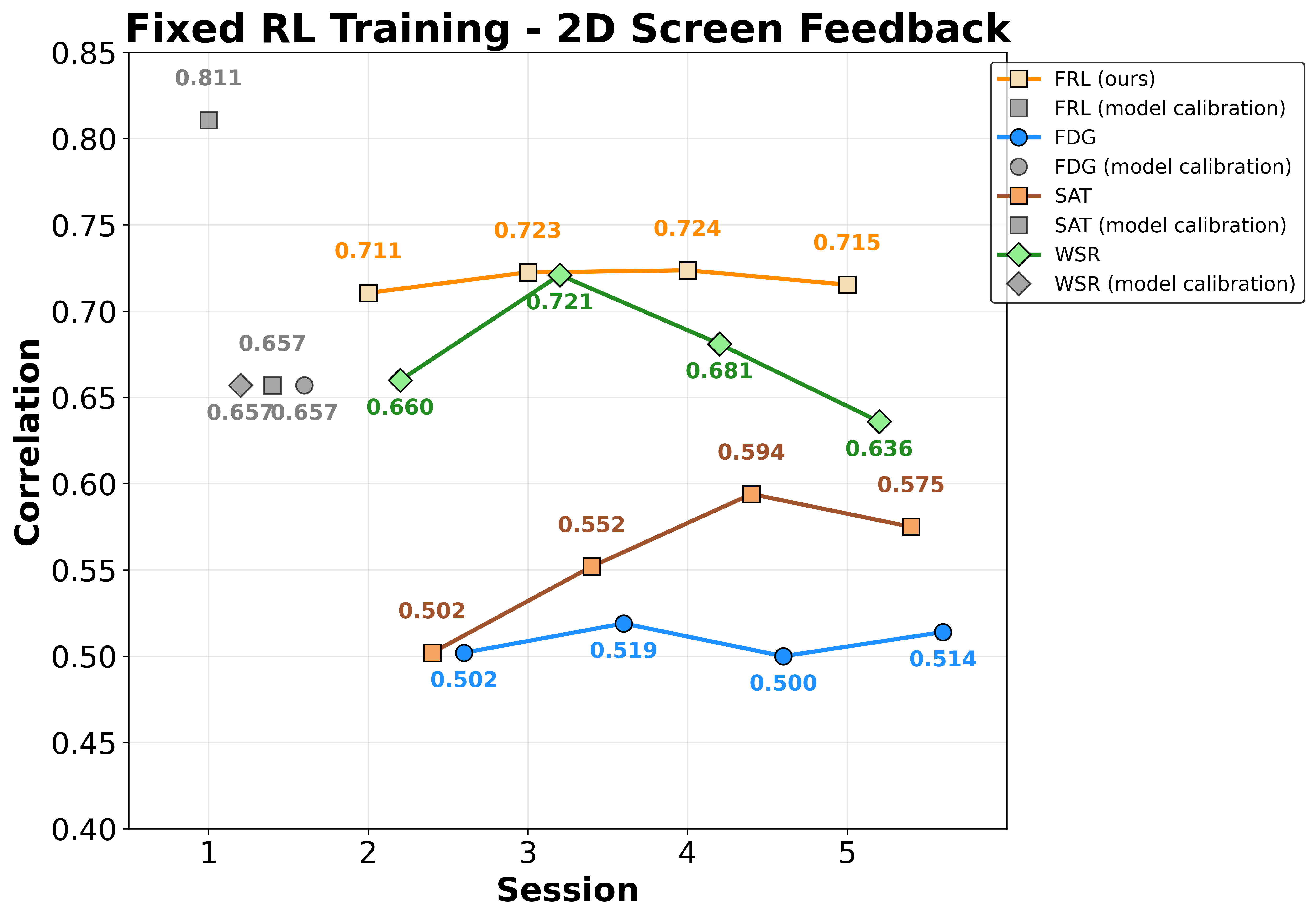}
        \caption{2D feedback}
        \label{fig:fold1}
    \end{subfigure}
    \hfill
    \begin{subfigure}[b]{0.48\linewidth}
        \centering
        \includegraphics[width=\linewidth]{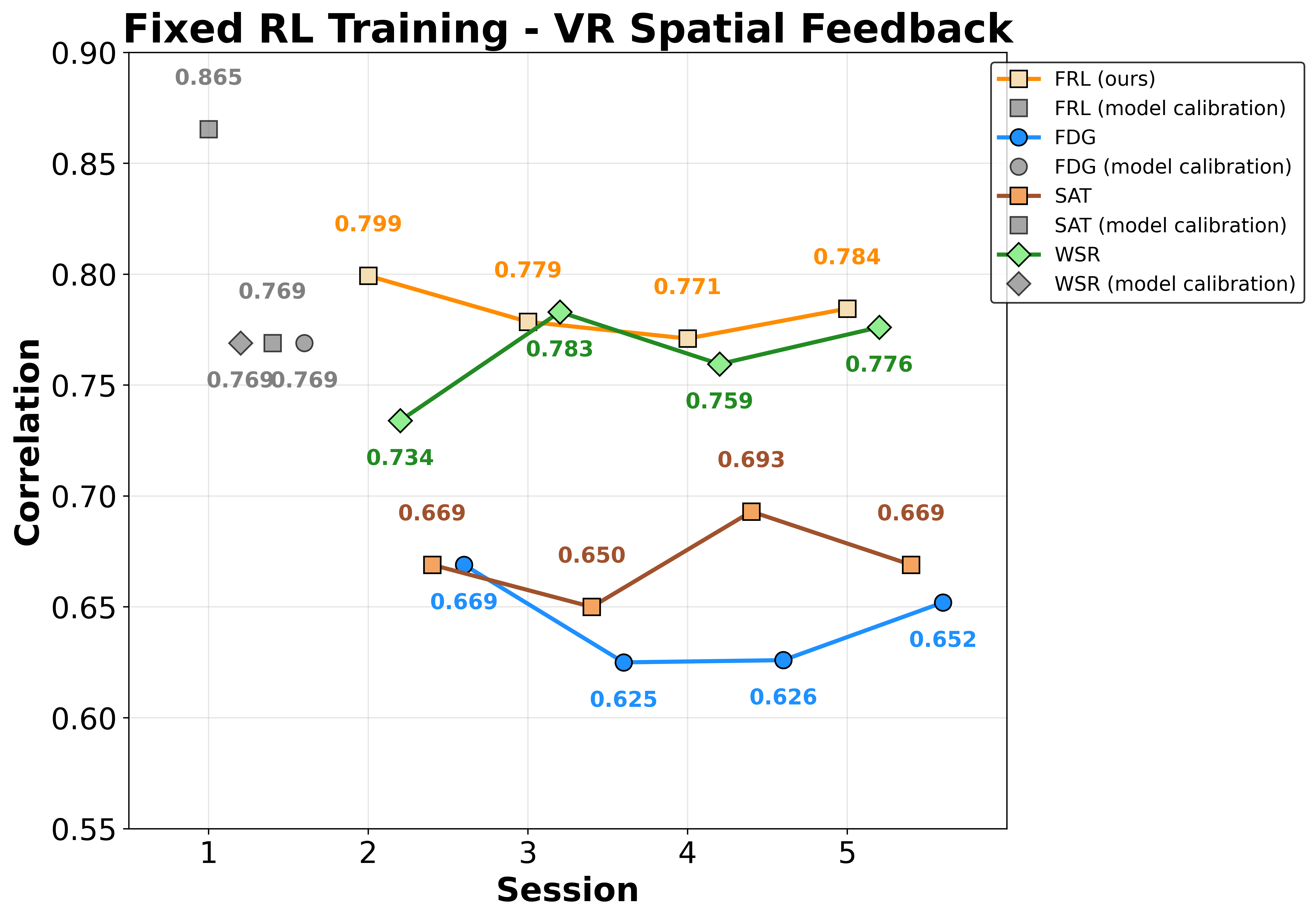}
        \caption{VR feedback}
        \label{fig:fold2}
    \end{subfigure}
    \caption{Pearson correlation coefficient of FRL comparison with three training strategies of BCI decoder in both 2D and VR feedback across sessions}
    \label{fig:comparesessions}
\end{figure}

Figure~\ref{fig:comparesessions} compares the trajectory correlation across five sessions using our method FRL with three BCI decoder strategies(FDG, SAT and WSR), under both 2D screen feedback and VR feedback. Each graph represents the mean correlation per session, while grey markers indicate performance when evaluated on the same session used for training (for FRL, FDG, and SAT). Under 2D feedback (Fig.~\ref{fig:comparesessions}a), FRL consistently achieves the highest correlation across all test sessions (Sessions 2--5), ranging from 0.711 to 0.724, with stable performance. This represents a substantial improvement over FDG (relative improvement of 39.1–44.8\%) and a clear advantage over SAT (up to 41.6\%). Interestingly, FRL even outperforms the upper bound WSR, with improvement from 0.28\% to 12.4\%. Under VR spatial feedback (Fig.~\ref{fig:comparesessions}b), FRL achieves the highest correlation, ranging from 0.771 to 0.799 across all sessions. This corresponds to clear improvements over FDG (19.4--24.6\%) and SAT (11.3--19.4\%). WSR performs closest to FRL and outperforms it in one session, but FRL maintains an overall advantage of 1.6--8.9\% across the remaining sessions.

Table \ref{tab:comparison_corr_all} illustrates that FRL achieves higher trajectory correlation than all compared strategies across both 2D and VR feedback on test sessions. Under 2D feedback, a one-way repeated measures ANOVA confirmed significant main effects among the strategies. FRL significantly outperforms FDG, with relative improvement of $41.5\%$. Similar significant improvements are observed over SAT ($29.1\%$). The relative gain over WSR ($6.2\%$) did not reach statistical significance. Under VR feedback, the Friedman test was applied and revealed significant differences in this group. The Wilcoxon signed-rank test and standardized Wilcoxon effect size $r_z$ were used. FRL maintains a clear advantage, significantly improving correlation over FDG with relative improvement of $21.2\%$ and over SAT ($20.6\%$). The minor gain over WSR ($2.5\%$) was not significant. 

\begin{table*}[t]
\caption{Comparison of Overall Pearson Correlation Coefficient: FRL vs. BCI Decoder Strategies under 2D and VR Feedback}
\label{tab:comparison_corr_all}
\centering
\small

\resizebox{\textwidth}{!}{
\renewcommand{\arraystretch}{1.4}
\setlength{\tabcolsep}{6pt} 

\begin{tabular}{l cccc c cc c cc} 
\toprule
\multicolumn{11}{c}{\textbf{2D Feedback}} \\
\toprule
\multirow{2}{*}{\textbf{Strategy}} & \multicolumn{4}{c}{\textbf{Correlation Mean (SD)}} & \multirow{2}{*}{$\Delta$} & \multicolumn{2}{c}{\textbf{$t$-test}} & \multirow{2}{*}{$|d|$} & \multicolumn{2}{c}{\textbf{ANOVA}} \\
\cmidrule(lr){2-5} \cmidrule(lr){7-8} \cmidrule(lr){10-11}
& X & Y & Z & Overall & & $|t(9)|$ & $p$ & & $F$ & $p$ \\
\midrule
FDG & 0.1209 (0.6885) & 0.7469 (0.3208) & 0.6551 (0.3678) & 0.5076 (0.3206) & +0.2105 & 5.8386 & 0.0005 & 1.8463 & \multirow{4}{*}{21.5892} & \multirow{4}{*}{$<0.0001$} \\
SAT & 0.1647 (0.6945) & 0.7894 (0.2719) & 0.7149 (0.3180) & 0.5563 (0.3029) & +0.1618 & 7.0388 & 0.0002 & 2.2259 & & \\
WSR & 0.3908 (0.6819) & 0.8416 (0.2332) & 0.7957 (0.2640) & 0.6761 (0.2984) & +0.0420 & 1.4620 & 0.1778 & 0.4600 & & \\
\textbf{FRL} & \textbf{0.2827 (0.6143)} & \textbf{0.9847 (0.0188)} & \textbf{0.8868 (0.1827)} & \textbf{0.7181 (0.2204)} & --- & --- & --- & --- & & \\
\midrule

\multicolumn{11}{c}{\textbf{VR Feedback}} \\
\toprule
\multirow{2}{*}{\textbf{Strategy}} & \multicolumn{4}{c}{\textbf{Correlation Mean (SD)}} & \multirow{2}{*}{$\Delta$} & \multicolumn{2}{c}{\textbf{Wilcoxon}} & \multirow{2}{*}{$r_z$} & \multicolumn{2}{c}{\textbf{Friedman}} \\
\cmidrule(lr){2-5} \cmidrule(lr){7-8} \cmidrule(lr){10-11}
& X & Y & Z & Overall & & $W$ & $p$ & & $Q$ & $p$ \\
\midrule
FDG & 0.2492 (0.6854) & 0.8455 (0.2243) & 0.8316 (0.2366) & 0.6420 (0.2870) & +0.1360 & 0.0 & 0.0059 & 0.8864 & \multirow{4}{*}{24.6} & \multirow{4}{*}{$<0.0001$} \\
SAT & 0.3019 (0.6727) & 0.8245 (0.2647) & 0.8090 (0.2689) & 0.6451 (0.2977) & +0.1329 & 0.0 & 0.0059 & 0.8864 & & \\
WSR & 0.5297 (0.6149) & 0.8797 (0.1829) & 0.8679 (0.1930) & 0.7591 (0.2557) & +0.0179 & 10.0 & 0.0840 & 0.5641 & & \\
\textbf{FRL} & \textbf{0.3749 (0.5449)} & \textbf{0.9838 (0.0184)} & \textbf{0.9754 (0.0322)} & \textbf{0.7780 (0.1825)} & --- & --- & --- & --- & & \\
\bottomrule

\multicolumn{11}{l}{\footnotesize * Note: $\Delta$ denotes the overall improvement of FRL; $p$ represent $p$-values. \quad $F$ is statistics value from one-way repeated measures ANOVA; $Q$ is from the Friedman test.} \\
\multicolumn{11}{l}{\footnotesize \quad $|t(9)|$ denotes absolute $t$-statistic; $W$ is the Wilcoxon statistic; $|d|$ and $r_z$ indicate Cohen's $d$ and standardized Wilcoxon effect size, respectively.}
\end{tabular}
}
\end{table*}

\subsection{Root Mean square Errors Comparison}
\begin{figure}[htbp]
    \centering
    \begin{subfigure}[b]{0.48\linewidth}
        \centering
        \includegraphics[width=\linewidth]{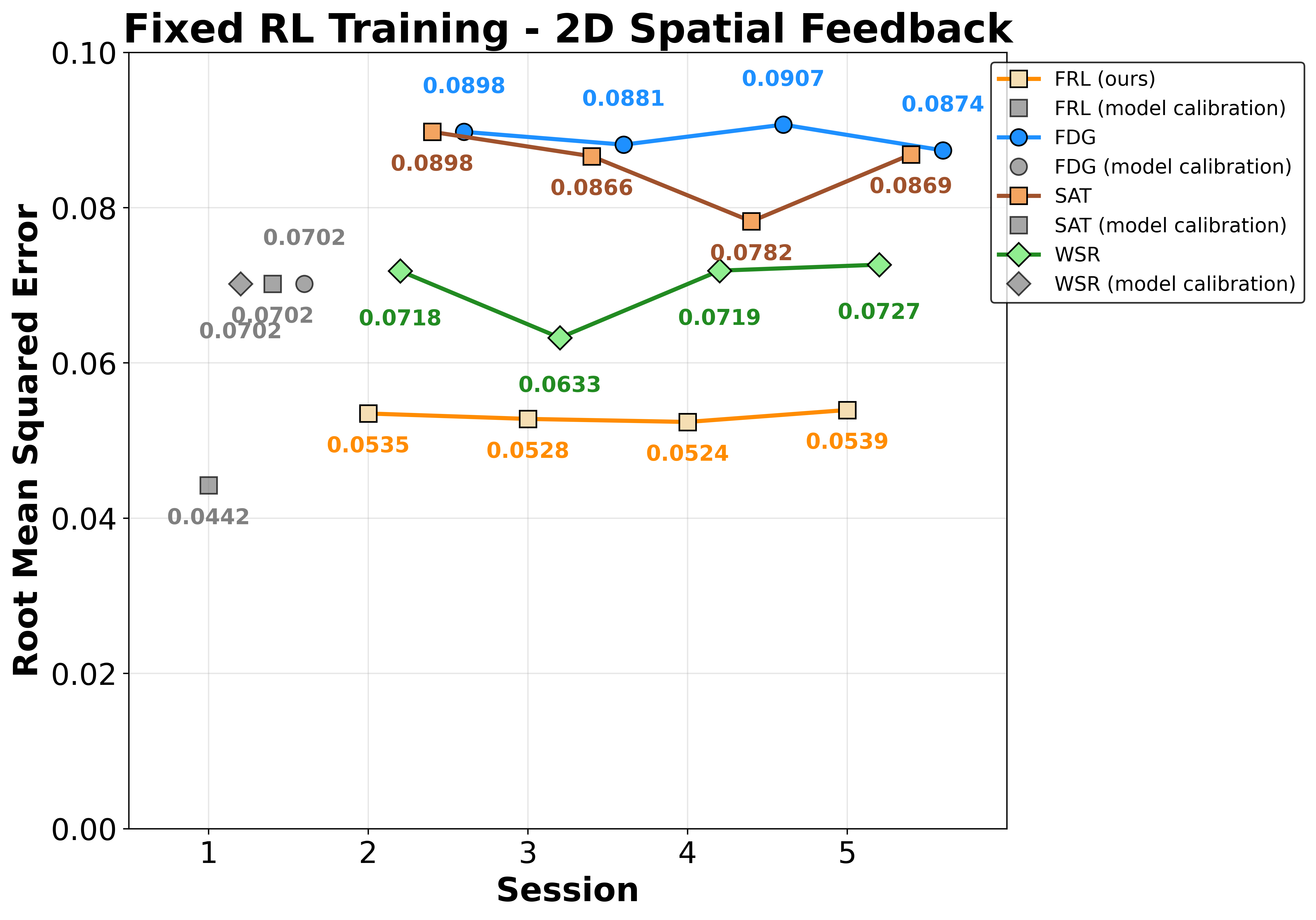}
        \caption{2D feedback}
    \end{subfigure}
    \hfill
    \begin{subfigure}[b]{0.48\linewidth}
        \centering
        \includegraphics[width=\linewidth]{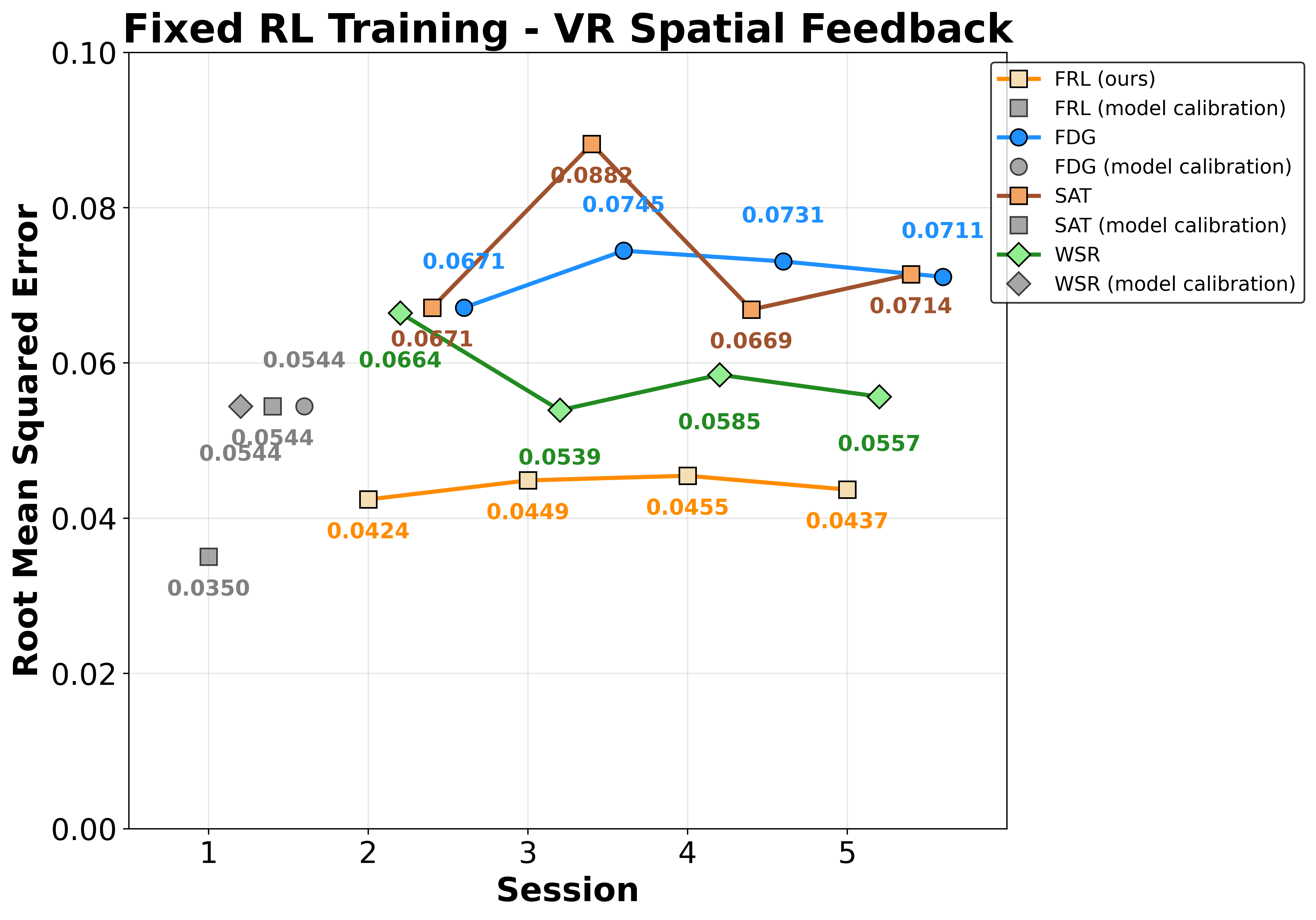}
        \caption{VR feedback}
    \end{subfigure}
    \caption{Root mean square errors comparison of FRL with three training strategies of BCI decoder in both 2D and VR feedback across sessions}
    \label{fig:comparesessions_RMSE}
\end{figure}

Figure~\ref{fig:comparesessions_RMSE} illustrates the RMSE across five sessions for FRL and three BCI decoder strategies (FDG, SAT, and WSR) under both 2D and VR feedback. Lower values indicate better performance. Under 2D feedback (Fig.~\ref{fig:comparesessions_RMSE}a), FRL consistently achieves the lowest RMSE (0.0524--0.0539), outperforming FDG (0.0874--0.0907, relative reduction of 38.3--42.2\%), SAT (0.0782--0.0898, 33.0--40.4\%), and WSR (0.0633--0.0727, 16.6--27.1\%).Under VR feedback (Fig.~\ref{fig:comparesessions_RMSE}b), FRL remains the most closely to the target (0.0424--0.0455), improving over FDG (0.0671--0.0745, 36.8--39.7\%), SAT (0.0669--0.0882, 32.0--49.1\%), and WSR (0.0539--0.0664, 16.7--36.1\%). 

\begin{table*}[t]
\caption{Comparison of Overall Root Mean Squared Error: FRL vs. BCI Decoder Strategies under 2D and VR Feedback}
\label{tab:comparison_RMSE_all}
\centering
\small

\resizebox{\textwidth}{!}{
\renewcommand{\arraystretch}{1.4}
\setlength{\tabcolsep}{6pt} 

\begin{tabular}{l cccc c cc c cc} 
\toprule
\multicolumn{11}{c}{\textbf{2D Feedback}} \\
\toprule
\multirow{2}{*}{\textbf{Strategy}} & \multicolumn{4}{c}{\textbf{RMSE Mean (SD)}} & \multirow{2}{*}{$\Delta$} & \multicolumn{2}{c}{\textbf{$t$-test}} & \multirow{2}{*}{$|d|$} & \multicolumn{2}{c}{\textbf{ANOVA}} \\
\cmidrule(lr){2-5} \cmidrule(lr){7-8} \cmidrule(lr){10-11}
& X & Y & Z & Overall & & $|t(9)|$ & $p$ & & $F$ & $p$ \\
\midrule
FDG & 0.1277 (0.0665) & 0.0649 (0.0377) & 0.0743 (0.0383) & 0.0890 (0.0335) & -0.0358 & 8.1108 & $<0.0001$ & 2.5649 & \multirow{4}{*}{38.8069} & \multirow{4}{*}{$<0.0001$} \\
SAT & 0.1239 (0.0673) & 0.0618 (0.0355) & 0.0704 (0.0358) & 0.0854 (0.0331) & -0.0322 & 10.9114 & $<0.0001$ & 3.4505 & & \\
WSR & 0.0985 (0.0705) & 0.0527 (0.0341) & 0.0585 (0.0346) & 0.0699 (0.0364) & -0.0167 & 4.2631 & 0.0021 & 1.3481 & & \\
\midrule
\textbf{FRL} & \textbf{0.1024 (0.0431)} & \textbf{0.0189 (0.0086)} & \textbf{0.0382 (0.0232)} & \textbf{0.0532 (0.0169)} & --- & --- & --- & --- & & \\
\midrule

\multicolumn{11}{c}{\textbf{VR Feedback}} \\
\toprule
\multirow{2}{*}{\textbf{Strategy}} & \multicolumn{4}{c}{\textbf{RMSE Mean (SD)}} & \multirow{2}{*}{$\Delta$} & \multicolumn{2}{c}{\textbf{$t$-test}} & \multirow{2}{*}{$|d|$} & \multicolumn{2}{c}{\textbf{ANOVA}} \\
\cmidrule(lr){2-5} \cmidrule(lr){7-8} \cmidrule(lr){10-11}
& X & Y & Z & Overall & & $|t(9)|$ & $p$ & & $F$ & $p$ \\
\midrule
FDG & 0.1139 (0.0676) & 0.0497 (0.0315) & 0.0507 (0.0313) & 0.0714 (0.0325) & -0.0273 & 14.1453 & $<0.0001$ & 4.4731 & \multirow{4}{*}{95.4544} & \multirow{4}{*}{$<0.0001$} \\
SAT & 0.1100 (0.6610) & 0.0541 (0.0360) & 0.0561 (0.0346) & 0.0734 (0.0343) & -0.0293 & 14.3031 & $<0.0001$ & 4.5230 & & \\
WSR & 0.0839 (0.0650) & 0.0452 (0.0294) & 0.0468 (0.0296) & 0.0586 (0.0327) & -0.0145 & 7.8100 & $<0.0001$ & 2.4687 & & \\
\midrule
\textbf{FRL} & \textbf{0.0929 (0.0518)} & \textbf{0.0182 (0.0088)} & \textbf{0.0213 (0.0122)} & \textbf{0.0441 (0.0186)} & --- & --- & --- & --- & & \\
\bottomrule

\multicolumn{11}{l}{\footnotesize * Note: $\Delta$ denotes the overall improvement of FRL; $p$ represent $p$-values. \quad $F$ is statistics value from one-way repeated measures ANOVA.} \\
\multicolumn{11}{l}{\footnotesize \quad $|t(9)|$ denotes absolute $t$-statistic; $W$ is the Wilcoxon statistic; $|d|$ and $r_z$ indicate Cohen's $d$ and standardized Wilcoxon effect size, respectively.}
\end{tabular}
}
\end{table*}

As shown in Table \ref{tab:comparison_RMSE_all}, under 2D feedback, FRL significantly reduces RMSE compared to FDG, decreasing the error from 0.0890 to 0.0532 (relative reduction of $40.2\%$). Similar significant improvements are observed over SAT ($37.7\%$) and WSR ($23.9\%$). Under VR feedback, FRL significantly reducing RMSE relative to FDG ($38.2\%$), SAT ($39.9\%$), and WSR ($24.7\%$).

\section{Discussion}
\subsection{Advantages of Residual Correction and Multi-objective Optimization}
The two-stage framework proposed in this paper significantly improves BCI decoding performance through an offline residual reinforcement learning approach. Specifically, FRL achieved its maximum performance gains with mean correlation improvements of 41.5\% (from 0.5076 to 0.7181) in 2D and 21.2\% (from 0.6420 to 0.7780) in VR, while simultaneously reducing RMSE by 40.2\% and 38.2\% in respective environments. The primary advantage of this residual architecture is its ability to compensate for suboptimal performance from fixed decoders, including limitations in training algorithms for decoding 3D motor imagery from non-stationary EEG, and variability introduced by user learning during 2D and VR feedback. By design, the correction process operates independently of direct EEG input and focuses exclusively on optimizing the kinematic output of the baseline decoder. Furthermore, the framework's decoupled design separates data collection from RL training without online policy updates that require continuous user interaction. Central to the system’s performance is a multi-objective optimization strategy that achieves synergy between global hyperparameter tuning and local kinematic rewards. By using Optuna\citep{optuna_2019} to maximize the Pearson correlation coefficient over 200 searching trials, the framework ensures that the decoded trajectories has similar morphology with the target trajectory. Simultaneously, the RL agent's reward function is designed to minimize the Euclidean distance between corrected and target velocities.

 The robustness of our approach is validated by rigorous statistical validation across all 10 participants. Similar to FDG, FRL is a strategy designed to assess a fixed decoder across sessions and thus we evaluate cross-session generality. FRL produced highly statistically significant improvements over both FDG and the periodic recalibration strategy, SAT, with large effect sizes ($|d|$ ranging from 1.8463 to 3.4505). Remarkably, in terms of RMSE, FRL outperformed WSR with high statistical significance. This indicates that RL effectively pushes the performance upper bound beyond what is attainable by CNN-LSTM BCI decoders alone through correcting systematic kinematic errors. Furthermore, FRL consistently exhibits a smaller range of performance fluctuations across the session timeline in both correlation and RMSE compared to all other decoding strategies. This reduced range across the extended experimental timeline indicates superior cross-session stability, proving that the FRL framework is highly resilient to the non-stationarity and signal drift inherent in longitudinal EEG data. Notably, in the VR environment, FRL consistently produces further significant enhancements, reaching large effect sizes in error reduction ($|d| = 4.4731$ to $4.5230$). Axis-specific analysis shows that FRL provides the largest improvements on the $Y$ and $Z$ axes, where it achieved near-perfect correlations (exceeding $0.97$ and reaching up to $0.9847$) and minimal RMSE (as low as $0.0182$ on the $Y$-axis). Both parametric and non-parametric tests converged to confirm that RL-based correction remains highly effective for error reduction even when baseline performance is strong. Given that previous CNN-LSTM architectures \citep{mc2023decoding} \citep{mcshane2026embodied} represented the existing performance benchmark for EEG-based 3D motion imagery, these results establish a new state-of-the-art.

\subsection{Limitations and Future Work}
The performance of the proposed residual reinforcement learning architecture is highly sensitive to the action scaling hyperparameter. A small action scale restricts the exploration space, resulting in limited performance improvement over the baseline BCI decoder. In contrast, a large action scale can lead to unstable corrections, causing the corrected velocity to deviate from the target. Because the residual framework explores locally around the baseline, its overall effectiveness is ultimately based upon the quality of the initial decoder. Additionally, the framework relies heavily on explicit temporal information. Without the observation of the time variable $t$, the agent's performance declines significantly. This dependency poses a challenge for real-time asynchronous BCI applications, where time-invariant decoding is essential for practical use; however in this experimental paradigm, cued home-to-target 3D motion imagery, synchronization of the RL is not issue. Despite the promising results, the current research is limited to a discrete four-target reaching task, which represents a relatively constrained experimental paradigm. In future work, we intend to expand the data collection protocol to include a broader range of motion trajectories and a higher density of target directions, as well as developing time invariant decoding. Additionally, while we have evaluated the cross-session generalisation capability of FRL through offline analysis, participant performance under real-time feedback, enhanced by the CNN-LSTM-RL framework, may further improve decoding accuracy and facilitate user learning. Consequently, real-time implementation and experimental validation suggest that this framework is likely to substantially advance 3D motor imagery.

\section{Conclusions}
We propose a two-stage framework to improve continuous 3D motor imagery decoding in non-invasive EEG-based brain–computer interfaces. First, a CNN–LSTM decoder is calibrated and used to collect synchronized EEG signals and target trajectories. Second, an offline residual reinforcement learning agent is trained to correct the fixed decoder’s outputs. We formulate a multi-objective optimization that minimizes Euclidean distance while maximizing Pearson correlation, with hyperparameters tuned via Optuna. Extensive experiments and statistical analysis show that the proposed method consistently outperforms baseline decoders in both correlation and RMSE under both 2D and VR environments. These results demonstrate that offline residual RL can effectively correct systematic decoding errors without additional neural input, providing a scalable solution for enhancing continuous 3D BCI control in applications such as virtual interaction, neurorehabilitation, and prosthetics.

\medskip

{
\small
\bibliographystyle{unsrtnat}
\bibliography{sample}

\begin{thebibliography}{48}
\providecommand{\natexlab}[1]{#1}
\providecommand{\url}[1]{\texttt{#1}}
\expandafter\ifx\csname urlstyle\endcsname\relax
  \providecommand{\doi}[1]{doi: #1}\else
  \providecommand{\doi}{doi: \begingroup \urlstyle{rm}\Url}\fi

\bibitem[Mill{\'a}n et~al.(2010)Mill{\'a}n, Rupp, M{\"u}ller-Putz, Murray-Smith, Giugliemma, Tangermann, Vidaurre, Cincotti, K{\"u}bler, Leeb, et~al.]{millan2010combining}
J~d~R Mill{\'a}n, R{\"u}diger Rupp, Gernot~R M{\"u}ller-Putz, Roderick Murray-Smith, Claudio Giugliemma, Michael Tangermann, Carmen Vidaurre, Febo Cincotti, Andrea K{\"u}bler, Robert Leeb, et~al.
\newblock Combining brain--computer interfaces and assistive technologies: state-of-the-art and challenges.
\newblock \emph{Frontiers in neuroscience}, 4:\penalty0 161, 2010.

\bibitem[Prasad et~al.(2010)Prasad, Herman, Coyle, McDonough, and Crosbie]{prasad2010applying}
Girijesh Prasad, Pawel Herman, Damien Coyle, Suzanne McDonough, and Jacqueline Crosbie.
\newblock Applying a brain-computer interface to support motor imagery practice in people with stroke for upper limb recovery: a feasibility study.
\newblock \emph{Journal of neuroengineering and rehabilitation}, 7\penalty0 (1):\penalty0 60, 2010.

\bibitem[Cervera et~al.(2018)Cervera, Soekadar, Ushiba, Mill{\'a}n, Liu, Birbaumer, and Garipelli]{cervera2018brain}
Mar{\'\i}a~A Cervera, Surjo~R Soekadar, Junichi Ushiba, Jos{\'e} del~R Mill{\'a}n, Meigen Liu, Niels Birbaumer, and Gangadhar Garipelli.
\newblock Brain-computer interfaces for post-stroke motor rehabilitation: a meta-analysis.
\newblock \emph{Annals of clinical and translational neurology}, 5\penalty0 (5):\penalty0 651--663, 2018.

\bibitem[Willett et~al.(2021)Willett, Avansino, Hochberg, Henderson, and Shenoy]{willett2021high}
Francis~R Willett, Donald~T Avansino, Leigh~R Hochberg, Jaimie~M Henderson, and Krishna~V Shenoy.
\newblock High-performance brain-to-text communication via handwriting.
\newblock \emph{Nature}, 593\penalty0 (7858):\penalty0 249--254, 2021.

\bibitem[Willett et~al.(2023)Willett, Kunz, Fan, Avansino, Wilson, Choi, Kamdar, Glasser, Hochberg, Druckmann, et~al.]{willett2023high}
Francis~R Willett, Erin~M Kunz, Chaofei Fan, Donald~T Avansino, Guy~H Wilson, Eun~Young Choi, Foram Kamdar, Matthew~F Glasser, Leigh~R Hochberg, Shaul Druckmann, et~al.
\newblock A high-performance speech neuroprosthesis.
\newblock \emph{Nature}, 620\penalty0 (7976):\penalty0 1031--1036, 2023.

\bibitem[Anumanchipalli et~al.(2019)Anumanchipalli, Chartier, and Chang]{anumanchipalli2019speech}
Gopala~K Anumanchipalli, Josh Chartier, and Edward~F Chang.
\newblock Speech synthesis from neural decoding of spoken sentences.
\newblock \emph{Nature}, 568\penalty0 (7753):\penalty0 493--498, 2019.

\bibitem[Du~Bois et~al.(2026)Du~Bois, Korik, Hodge, Hudson, Elahi, Bigirimana, Dayan, Sanchez-Bornot, McCann, Yelden, et~al.]{du2026advancing}
Naomi Du~Bois, Attila Korik, Stephanie Hodge, Leah Hudson, Ainjila~S Elahi, Alain Bigirimana, Natalie Dayan, Jose~M Sanchez-Bornot, Alison McCann, Kudret Yelden, et~al.
\newblock Advancing eeg-based assessment of consciousness and cognition in prolonged disorders of consciousness.
\newblock \emph{Communications medicine}, 2026.

\bibitem[Marshall et~al.(2013)Marshall, Coyle, Wilson, and Callaghan]{marshall2013games}
David Marshall, Damien Coyle, Shane Wilson, and Michael Callaghan.
\newblock Games, gameplay, and bci: the state of the art.
\newblock \emph{IEEE Transactions on Computational Intelligence and AI in Games}, 5\penalty0 (2):\penalty0 82--99, 2013.

\bibitem[van~de Laar et~al.(2013)van~de Laar, G{\"u}rk{\"o}k, Plass-Oude~Bos, Poel, and Nijholt]{van2013experiencing}
Bram van~de Laar, Hayrettin G{\"u}rk{\"o}k, Danny Plass-Oude~Bos, Mannes Poel, and Anton Nijholt.
\newblock Experiencing bci control in a popular computer game.
\newblock \emph{IEEE Transactions on Computational Intelligence and AI in Games}, 5\penalty0 (2):\penalty0 176--184, 2013.

\bibitem[Hochberg et~al.(2012)Hochberg, Bacher, Jarosiewicz, Masse, Simeral, Vogel, Haddadin, Liu, Cash, Van Der~Smagt, et~al.]{hochberg2012reach}
Leigh~R Hochberg, Daniel Bacher, Beata Jarosiewicz, Nicolas~Y Masse, John~D Simeral, Joern Vogel, Sami Haddadin, Jie Liu, Sydney~S Cash, Patrick Van Der~Smagt, et~al.
\newblock Reach and grasp by people with tetraplegia using a neurally controlled robotic arm.
\newblock \emph{Nature}, 485\penalty0 (7398):\penalty0 372--375, 2012.

\bibitem[Ajiboye et~al.(2017)Ajiboye, Willett, Young, Memberg, Murphy, Miller, Walter, Sweet, Hoyen, Keith, et~al.]{ajiboye2017restoration}
A~Bolu Ajiboye, Francis~R Willett, Daniel~R Young, William~D Memberg, Brian~A Murphy, Jonathan~P Miller, Benjamin~L Walter, Jennifer~A Sweet, Harry~A Hoyen, Michael~W Keith, et~al.
\newblock Restoration of reaching and grasping movements through brain-controlled muscle stimulation in a person with tetraplegia: a proof-of-concept demonstration.
\newblock \emph{The Lancet}, 389\penalty0 (10081):\penalty0 1821--1830, 2017.

\bibitem[Edelman et~al.(2019)Edelman, Meng, Suma, Zurn, Nagarajan, Baxter, Cline, and He]{edelman2019noninvasive}
Bradley~J Edelman, Jianjun Meng, Daniel Suma, Claire Zurn, Eric Nagarajan, Bryan~S Baxter, Christopher~C Cline, and BJSR He.
\newblock Noninvasive neuroimaging enhances continuous neural tracking for robotic device control.
\newblock \emph{Science robotics}, 4\penalty0 (31):\penalty0 eaaw6844, 2019.

\bibitem[Lebedev and Nicolelis(2006)]{lebedev2006brain}
Mikhail~A Lebedev and Miguel~AL Nicolelis.
\newblock Brain--machine interfaces: past, present and future.
\newblock \emph{TRENDS in Neurosciences}, 29\penalty0 (9):\penalty0 536--546, 2006.

\bibitem[Lotte et~al.(2018)Lotte, Bougrain, Cichocki, Clerc, Congedo, Rakotomamonjy, and Yger]{lotte2018review}
Fabien Lotte, Laurent Bougrain, Andrzej Cichocki, Maureen Clerc, Marco Congedo, Alain Rakotomamonjy, and Florian Yger.
\newblock A review of classification algorithms for eeg-based brain--computer interfaces: a 10 year update.
\newblock \emph{Journal of neural engineering}, 15\penalty0 (3):\penalty0 031005, 2018.

\bibitem[Abiri et~al.(2019)Abiri, Borhani, Sellers, Jiang, and Zhao]{abiri2019comprehensive}
Reza Abiri, Soheil Borhani, Eric~W Sellers, Yang Jiang, and Xiaopeng Zhao.
\newblock A comprehensive review of eeg-based brain--computer interface paradigms.
\newblock \emph{Journal of neural engineering}, 16\penalty0 (1):\penalty0 011001, 2019.

\bibitem[Bradberry et~al.(2010)Bradberry, Gentili, and Contreras-Vidal]{bradberry2010reconstructing}
Trent~J Bradberry, Rodolphe~J Gentili, and Jos{\'e}~L Contreras-Vidal.
\newblock Reconstructing three-dimensional hand movements from noninvasive electroencephalographic signals.
\newblock \emph{Journal of neuroscience}, 30\penalty0 (9):\penalty0 3432--3437, 2010.

\bibitem[Chen et~al.(2022)Chen, Fu, Wu, Song, Ma, Jiang, and Zhang]{chen2022continuous}
Yi-Feng Chen, Ruiqi Fu, Junde Wu, Jongbin Song, Rui Ma, Yi-Chuan Jiang, and Mingming Zhang.
\newblock Continuous bimanual trajectory decoding of coordinated movement from eeg signals.
\newblock \emph{IEEE Journal of Biomedical and Health Informatics}, 26\penalty0 (12):\penalty0 6012--6023, 2022.

\bibitem[Wang et~al.(2023)Wang, Cao, Zhou, Gong, Yousefnezhad, Shao, and Zhang]{wang2023comprehensive}
Pengpai Wang, Xuhao Cao, Yueying Zhou, Peiliang Gong, Muhammad Yousefnezhad, Wei Shao, and Daoqiang Zhang.
\newblock A comprehensive review on motion trajectory reconstruction for eeg-based brain-computer interface.
\newblock \emph{Frontiers in neuroscience}, 17:\penalty0 1086472, 2023.

\bibitem[Xu et~al.(2021{\natexlab{a}})Xu, Wang, Deng, Wu, Zhang, Li, and Song]{xu2021decoding}
Baoguo Xu, Yong Wang, Leying Deng, Changcheng Wu, Wenbing Zhang, Huijun Li, and Aiguo Song.
\newblock Decoding hand movement types and kinematic information from electroencephalogram.
\newblock \emph{IEEE Transactions on Neural Systems and Rehabilitation Engineering}, 29:\penalty0 1744--1755, 2021{\natexlab{a}}.

\bibitem[Korik et~al.(2025)Korik, Du~Bois, Bornot, McShane, Guger, Del~Felice, Lennon, and Coyle]{korik2025decoding}
Attila Korik, Naomi Du~Bois, Jose~Sanchez Bornot, Niall McShane, Christoph Guger, Alessandra Del~Felice, Olive Lennon, and Damien Coyle.
\newblock Decoding the variable velocity of lower-limb stepping movements from eeg.
\newblock \emph{IEEE Transactions on Neural Systems and Rehabilitation Engineering}, 2025.

\bibitem[Mc~Shane et~al.(2023)Mc~Shane, McCreadie, Charles, Korik, and Coyle]{mc2023decoding}
Niall Mc~Shane, Karl McCreadie, Darryl Charles, Attila Korik, and Damien Coyle.
\newblock Decoding motion trajectories in an upper limb bci: Linear regression vs deep learning.
\newblock In \emph{2023 IEEE International Conference on Metrology for eXtended Reality, Artificial Intelligence and Neural Engineering (MetroXRAINE)}, pages 1039--1044. IEEE, 2023.

\bibitem[McShane et~al.(2026)McShane, Korik, McCreadie, Bois, Charles, and Coyle]{mcshane2026embodied}
Niall McShane, Attila Korik, Karl McCreadie, Naomi~Du Bois, Darryl Charles, and Damien Coyle.
\newblock Embodied virtual reality feedback reshapes neural representations to support continuous three-dimensional motor imagery decoding.
\newblock \emph{arXiv preprint arXiv:2605.29677}, 2026.

\bibitem[Mc~Shane et~al.(2022)Mc~Shane, McCreadie, Charles, Korik, and Coyle]{mc2022online}
Niall Mc~Shane, Karl McCreadie, Darryl Charles, Attila Korik, and Damien Coyle.
\newblock Online 3d motion decoder bci for embodied virtual reality upper limb control: A pilot study.
\newblock In \emph{2022 IEEE International Conference on Metrology for Extended Reality, Artificial Intelligence and Neural Engineering (Metroxraine)}, pages 697--702. IEEE, 2022.

\bibitem[Sutton et~al.(1998)Sutton, Barto, et~al.]{sutton1998reinforcement}
Richard~S Sutton, Andrew~G Barto, et~al.
\newblock \emph{Reinforcement learning: An introduction}, volume~1.
\newblock MIT press Cambridge, 1998.

\bibitem[Kober et~al.(2013)Kober, Bagnell, and Peters]{kober2013reinforcement}
Jens Kober, J~Andrew Bagnell, and Jan Peters.
\newblock Reinforcement learning in robotics: A survey.
\newblock \emph{The International Journal of Robotics Research}, 32\penalty0 (11):\penalty0 1238--1274, 2013.

\bibitem[Singh et~al.(2022)Singh, Kumar, and Singh]{singh2022reinforcement}
Bharat Singh, Rajesh Kumar, and Vinay~Pratap Singh.
\newblock Reinforcement learning in robotic applications: a comprehensive survey.
\newblock \emph{Artificial intelligence review}, 55\penalty0 (2):\penalty0 945--990, 2022.

\bibitem[Tang et~al.(2025)Tang, Abbatematteo, Hu, Chandra, Mart{\'\i}n-Mart{\'\i}n, and Stone]{tang2025deep}
Chen Tang, Ben Abbatematteo, Jiaheng Hu, Rohan Chandra, Roberto Mart{\'\i}n-Mart{\'\i}n, and Peter Stone.
\newblock Deep reinforcement learning for robotics: A survey of real-world successes.
\newblock \emph{Annual Review of Control, Robotics, and Autonomous Systems}, 8\penalty0 (1):\penalty0 153--188, 2025.

\bibitem[Sharifi et~al.(2025)Sharifi, Tripathi, Chen, Zhang, and Tavakoli]{sharifi2025reinforcement}
Mojtaba Sharifi, Shreesh Tripathi, Yun Chen, Qiang Zhang, and Mahdi Tavakoli.
\newblock Reinforcement learning methods for assistive and rehabilitation robotic systems: A survey.
\newblock \emph{IEEE Transactions on Systems, Man, and Cybernetics: Systems}, 2025.

\bibitem[Kim et~al.(2017)Kim, Kirchner, Stefes, and Kirchner]{kim2017intrinsic}
Su~Kyoung Kim, Elsa~Andrea Kirchner, Arne Stefes, and Frank Kirchner.
\newblock Intrinsic interactive reinforcement learning--using error-related potentials for real world human-robot interaction.
\newblock \emph{Scientific reports}, 7\penalty0 (1):\penalty0 17562, 2017.

\bibitem[Wang et~al.(2022)Wang, Chen, and Lin]{wang2022error}
Xiaofei Wang, Hsiang-Ting Chen, and Chin-Teng Lin.
\newblock Error-related potential-based shared autonomy via deep recurrent reinforcement learning.
\newblock \emph{Journal of Neural Engineering}, 19\penalty0 (6):\penalty0 066023, 2022.

\bibitem[Xavier~Fid{\^e}ncio et~al.(2025)Xavier~Fid{\^e}ncio, Gr{\"u}n, Klaes, and Iossifidis]{xavier2025hybrid}
Aline Xavier~Fid{\^e}ncio, Felix Gr{\"u}n, Christian Klaes, and Ioannis Iossifidis.
\newblock Hybrid brain-computer interface using error-related potential and reinforcement learning.
\newblock \emph{Frontiers in Human Neuroscience}, 19:\penalty0 1569411, 2025.

\bibitem[Xu et~al.(2026)Xu, Bi, Yang, Ge, Wang, Lian, Fei, and Zhang]{xu2026brain}
Zichao Xu, Luzheng Bi, Zhenge Yang, Haorui Ge, Zitong Wang, Kaixuan Lian, Weijie Fei, and Peiyu Zhang.
\newblock Brain-controlled operator model-driven deep reinforcement learning for adaptive brain-machine collaborative control.
\newblock \emph{Expert Systems with Applications}, 305:\penalty0 130770, 2026.

\bibitem[Agarwal et~al.(2020)Agarwal, Venkateswaran, and Sivakumar]{agarwal2020human}
Mohit Agarwal, Shyam~Krishnan Venkateswaran, and Raghupathy Sivakumar.
\newblock Human-in-the-loop rl with an eeg wearable headset: On effective use of brainwaves to accelerate learning.
\newblock In \emph{Proceedings of the 6th ACM Workshop on Wearable Systems and Applications}, pages 25--30, 2020.

\bibitem[Xu et~al.(2021{\natexlab{b}})Xu, Agarwal, Gupta, Fekri, and Sivakumar]{xu2021accelerating}
Duo Xu, Mohit Agarwal, Ekansh Gupta, Faramarz Fekri, and Raghupathy Sivakumar.
\newblock Accelerating reinforcement learning using eeg-based implicit human feedback.
\newblock \emph{Neurocomputing}, 460:\penalty0 139--153, 2021{\natexlab{b}}.

\bibitem[Zhu et~al.(2025)Zhu, Wang, and Gao]{zhu2025eeg}
Hao Zhu, Jialin Wang, and Rui Gao.
\newblock Eeg-based inverse reinforcement learning for safety-oriented global path planning in dynamic environments.
\newblock \emph{Applied Sciences}, 15\penalty0 (11):\penalty0 6163, 2025.

\bibitem[Silver et~al.(2018)Silver, Allen, Tenenbaum, and Kaelbling]{silver2018residual}
Tom Silver, Kelsey Allen, Josh Tenenbaum, and Leslie Kaelbling.
\newblock Residual policy learning.
\newblock \emph{arXiv preprint arXiv:1812.06298}, 2018.

\bibitem[Johannink et~al.(2019)Johannink, Bahl, Nair, Luo, Kumar, Loskyll, Ojea, Solowjow, and Levine]{johannink2019residual}
Tobias Johannink, Shikhar Bahl, Ashvin Nair, Jianlan Luo, Avinash Kumar, Matthias Loskyll, Juan~Aparicio Ojea, Eugen Solowjow, and Sergey Levine.
\newblock Residual reinforcement learning for robot control.
\newblock In \emph{2019 international conference on robotics and automation (ICRA)}, pages 6023--6029. IEEE, 2019.

\bibitem[Alakuijala et~al.(2021)Alakuijala, Dulac-Arnold, Mairal, Ponce, and Schmid]{alakuijala2021residual}
Minttu Alakuijala, Gabriel Dulac-Arnold, Julien Mairal, Jean Ponce, and Cordelia Schmid.
\newblock Residual reinforcement learning from demonstrations.
\newblock \emph{arXiv preprint arXiv:2106.08050}, 2021.

\bibitem[Davchev et~al.(2022)Davchev, Luck, Burke, Meier, Schaal, and Ramamoorthy]{davchev2022residual}
Todor Davchev, Kevin~Sebastian Luck, Michael Burke, Franziska Meier, Stefan Schaal, and Subramanian Ramamoorthy.
\newblock Residual learning from demonstration: Adapting dmps for contact-rich manipulation.
\newblock \emph{IEEE Robotics and Automation Letters}, 7\penalty0 (2):\penalty0 4488--4495, 2022.

\bibitem[Hameed et~al.(2025)Hameed, Khan, Ahmed, Aftab, and Fazal]{hameed2025enhancing}
Ibtehaaj Hameed, Danish~M Khan, Syed~Muneeb Ahmed, Syed~Sabeeh Aftab, and Hammad Fazal.
\newblock Enhancing motor imagery eeg signal decoding through machine learning: A systematic review of recent progress.
\newblock \emph{Computers in Biology and Medicine}, 185:\penalty0 109534, 2025.

\bibitem[Makeig(1993)]{makeig1993auditory}
Scott Makeig.
\newblock Auditory event-related dynamics of the eeg spectrum and effects of exposure to tones.
\newblock \emph{Electroencephalography and clinical neurophysiology}, 86\penalty0 (4):\penalty0 283--293, 1993.

\bibitem[Haarnoja et~al.(2018)Haarnoja, Zhou, Hartikainen, Tucker, Ha, Tan, Kumar, Zhu, Gupta, Abbeel, et~al.]{haarnoja2018soft}
Tuomas Haarnoja, Aurick Zhou, Kristian Hartikainen, George Tucker, Sehoon Ha, Jie Tan, Vikash Kumar, Henry Zhu, Abhishek Gupta, Pieter Abbeel, et~al.
\newblock Soft actor-critic algorithms and applications.
\newblock \emph{arXiv preprint arXiv:1812.05905}, 2018.

\bibitem[Akiba et~al.(2019)Akiba, Sano, Yanase, Ohta, and Koyama]{optuna_2019}
Takuya Akiba, Shotaro Sano, Toshihiko Yanase, Takeru Ohta, and Masanori Koyama.
\newblock Optuna: A next-generation hyperparameter optimization framework.
\newblock In \emph{Proceedings of the 25th {ACM} {SIGKDD} International Conference on Knowledge Discovery and Data Mining}, 2019.

\bibitem[Girden(1992)]{girden1992anova}
Ellen~R Girden.
\newblock \emph{ANOVA: Repeated measures}.
\newblock Number~84. sage, 1992.

\bibitem[Friedman(1937)]{friedman1937use}
Milton Friedman.
\newblock The use of ranks to avoid the assumption of normality implicit in the analysis of variance.
\newblock \emph{Journal of the american statistical association}, 32\penalty0 (200):\penalty0 675--701, 1937.

\bibitem[Wilcoxon(1992)]{wilcoxon1992individual}
Frank Wilcoxon.
\newblock Individual comparisons by ranking methods.
\newblock In \emph{Breakthroughs in statistics: Methodology and distribution}, pages 196--202. Springer, 1992.

\bibitem[Cohen(2013)]{cohen2013statistical}
Jacob Cohen.
\newblock \emph{Statistical power analysis for the behavioral sciences}.
\newblock routledge, 2013.

\bibitem[Holm(1979)]{holm1979simple}
Sture Holm.
\newblock A simple sequentially rejective multiple test procedure.
\newblock \emph{Scandinavian journal of statistics}, pages 65--70, 1979.

\end{thebibliography}

}


\newpage
\begin{appendices}
\section{Ethics Statement}
\label{app:ethics}
This study received ethical approval from the Institutional Research Ethics Committee of the authors' Institutes in accordance with the Declaration of Helsinki and institutional guidelines for research involving human subjects. Participants provided written informed consent prior to participation and were informed of the study procedures, data recording methods (including EEG, kinematic and eye-tracking data), their right to withdraw at any time without penalty, and the intended research use of anonymised data. All data were anonymised prior to analysis and dissemination. No personally identifiable information is included in this publication.
\section{Reach Targets}
\label{app:reach targets}
\begin{center}
    \includegraphics[width=1.0\textwidth]{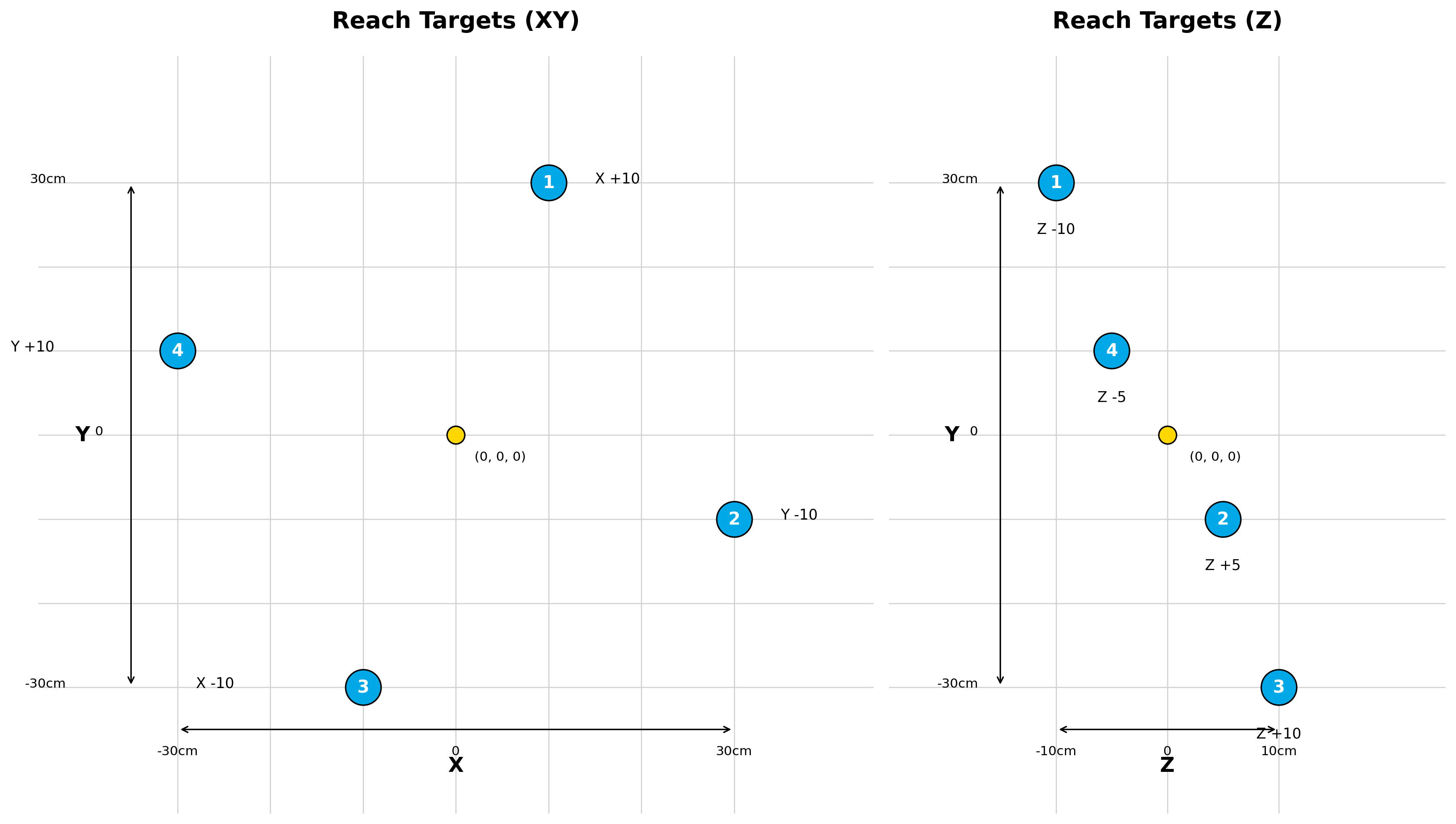}
\end{center}

\section{Hyperparameters Searching}
\label{app:hyperparameters searching}
\begin{table}[H]
\centering
\caption{Hyperparameter search space for the RL agent}
\label{tab:hyperparameter_search}
\begin{tabular}{lll}
\hline
\textbf{Parameter} & \textbf{Type} & \textbf{Search Range} \\
\hline
Actor learning rate    & Log-uniform & $[10^{-5}, 10^{-3}]$ \\
Critic learning rate  & Log-uniform & $[10^{-5}, 10^{-3}]$ \\
Hidden layer dimension          & Categorical & $\{128, 256, 512\}$ \\
Batch size                                           & Categorical & $\{64, 128, 256\}$ \\
Reward Discount factor                             & Uniform     & $[0.90, 0.999]$ \\
SAC Target smoothing coefficient                  & Uniform     & $[0.005, 0.05]$ \\
Bias scale (x-axis) ($Scale_1$)                           & Uniform     & $[0.1, 1.0]$ \\
Bias scale (y-axis) ($Scale_2$)                           & Uniform     & $[0.1, 1.0]$ \\
Bias scale (z-axis) ($Scale_3$)                           & Uniform     & $[0.1, 1.0]$ \\
Number of actor layers                                & Integer     & $[2, 4]$ \\
Number of critic layers                               & Integer     & $[2, 4]$ \\
\hline
\end{tabular}
\end{table}
\section{Prerequisite Analysis for Statistical Tests}
\label{app:statistical_analysis}

As an additional repeated-measures modelling check, linear mixed-effects models were explored separately for the 2D and VR feedback conditions, with decoder strategy specified as a fixed effect and participant specified as a random intercept. However, these models produced singular fits in one or more metric-by-feedback analyses, indicating that the participant-level random-intercept variance was estimated at or near zero and that the random-effects structure was not reliably supported by the available participant-level data. Therefore, LMM-based inference was not retained as the primary statistical framework. Instead, test selection was based on the distributional diagnostics within each feedback condition: paired t-tests were used where paired differences satisfied normality and symmetry assumptions, while Wilcoxon signed-rank tests were used where these assumptions were not met or where a consistent non-parametric approach was required within a metric. The following statistical analyses serve as a prerequisite validation for determining the appropriate hypothesis tests. Specifically, Shapiro-Wilk tests and Quantile-Quantile plots (Q-Q plots) are employed to assess the normality of the data distribution, while histograms and skewness coefficients are used to evaluate symmetry. These diagnostics were used to determine whether parametric paired comparisons were appropriate, or whether non-parametric alternatives were required, for a statistical comparison (correlation and RMSE) between reinforcement learning FRL and BCI decoder strategies (FDG, SAT and WSR).

\subsection{With 2D feedback}

The prerequisite analyses for the 2D feedback condition demonstrate that all performance metrics for the compared BCI decoder strategies (FDG, SAT, and WSR vs. FRL) satisfy the normality assumption. The data points in Q-Q plots for both Pearson correlation and RMSE differences align closely with the diagonal reference lines, which is further supported by Shapiro-Wilk test results where all $p$-values are greater than 0.05. Additionally, histograms and skewness coefficients confirm that these distributions are relatively symmetric. Given these characteristics, the paired t-tests were therefore used for the 2D correlation and RMSE pairwise comparisons.
\subsection{With VR feedback}
In the VR spatial feedback environment, Q-Q plots and Shapiro-Wilk tests generally indicated that the majority of comparisons followed a normal distribution ($p > 0.05$). However, the Pearson correlation differences for the FDG vs. FRL comparison proved to be an exception, deviating from the reference line in its Q-Q plot and yielding a Shapiro-Wilk $p$-value of $0.0136$. To maintain statistical consistency across this metric, we employed the Wilcoxon Signed-Rank test for the entire set of correlation comparisons (FRL vs. FDG, SAT, and WSR). This non-parametric approach was deemed appropriate as the distribution of FDG correlation differences remained sufficiently symmetric ($\text{skewness} = -0.2609$). In contrast, the RMSE comparison group satisfied the normality assumption across all pairs; thus, Paired $t$-tests were utilized to assess those differences.





\section{Experiments Compute Resources}
\label{app:compute resources}
All experiments were conducted using NVIDIA RTX 3090 GPUs. The RL agents were implemented in PyTorch and trained with CUDA acceleration. Each experiment uses 8 GPUs and requires approximately 24 hours of wall-clock time for hyperparameter optimization, corresponding to 192 GPU hours per run. A total of 20 experiments were conducted, resulting in approximately 3840 GPU hours for hyperparameter optimization. In addition, the final training of RL agents for each participant and feedback condition requires approximately 3 GPU hours per run, leading to an additional cost of 60 GPU hours. Overall, the total computational cost is approximately 3900 GPU hours. Experiments were executed in parallel with up to 6 runs simultaneously, resulting in a total wall-clock time of approximately 6 days.

\end{appendices}



\end{document}